\documentclass[11pt]{article}

\usepackage[margin=1in]{geometry}

\usepackage{amsmath}
\usepackage{amssymb}
\usepackage{amsfonts}
\usepackage{amsthm}
\usepackage{mathtools}

\usepackage{graphicx}
\usepackage{booktabs}
\usepackage{multirow}
\usepackage{subcaption}

\usepackage{enumitem}
\usepackage{xcolor}

\usepackage{cite}

\usepackage{url}
\usepackage{hyperref}

\usepackage{microtype}

\usepackage{bm}

\hypersetup{
    colorlinks=true,
    linkcolor=blue,
    citecolor=blue,
    urlcolor=blue
}

\newtheorem{theorem}{Theorem}[section]
\newtheorem{lemma}[theorem]{Lemma}

\theoremstyle{definition}
\newtheorem{definition}[theorem]{Definition}

\theoremstyle{remark}
\newtheorem{remark}[theorem]{Remark}

\title{
    \textbf{Improving Performance of Spike-based Deep Q-Learning
    using Ternary Neurons}
}

\author{
Aref Ghoreishee,
Abhishek Mishra,
John Walsh,
Anup Das,
and Nagarajan Kandasamy
\\[6pt]
\small Department of Electrical and Computer Engineering\\
\small Drexel University\\
\small Philadelphia, PA, USA
\\[4pt]
\small
\texttt{
\{ag4247, am48626, jmw96, ad3639, kandasamy\}@drexel.edu
}
}

\date{}

\begin{document}

\maketitle


\begin{abstract}
We propose a new ternary spiking neuron model to improve the representation capacity of binary spiking neurons in deep Q-learning. Although a ternary neuron model has recently been introduced to overcome the limited representation capacity offered by binary spiking neurons, we show that its performance is worse than that of binary models in deep Q-learning tasks, contradicting previous findings from recent studies. Through mathematical and empirical analysis, we hypothesize that gradient estimation bias during training is the underlying cause. The proposed ternary spiking neuron model mitigates this issue by reducing the estimation bias. We use the proposed ternary spiking neuron as the fundamental computing unit in a deep spiking Q-learning network, which we call the deep asymmetric ternary spiking Q-network (DATSQN), and evaluate the network's performance in seven Atari games from the Gym environment. The results show that the proposed ternary spiking neuron mitigates the performance degradation of ternary neurons in DQN tasks and improves the mean game score relative to the binary baseline under the evaluation settings used in this paper.
\end{abstract}

\vspace{0.5em}



\section{Introduction}

\emph{Deep reinforcement learning} (\emph{DRL}) has demonstrated strong performance in decision making across numerous application domains, including autonomous vehicle and robot navigation in dynamic environments~\cite{hu2021sim}. However, the power consumed by DRL-based algorithms, especially those built on classical deep neural networks (DNNs) such as deep Q-networks (DQN), during inference can limit their deployment on mobile platforms. \emph{Spiking neural networks} (\emph{SNNs}) offer an energy-efficient alternative. These networks are inspired by the operational principles of biological neurons, which communicate by sending short impulses, called spikes, through synapses. Such spiking neurons can be organized into feed-forward layers or recurrent topologies. This brain-inspired computing model is well suited to spatial and temporal pattern recognition tasks.



The inherent computational efficiency of SNNs has motivated the development of deep spiking reinforcement learning networks such as \emph{deep spiking Q-networks} (\emph{DSQN}), in which classical neurons are replaced by leaky integrate-and-fire (LIF) spiking neurons that produce binary outputs (0 or 1) over a predefined simulation time window~\cite{feng2024brainqn,tang2021deep}. However, SNNs currently lag behind their conventional counterparts in performance for two main reasons.

First, the non-differentiable nature of spiking neurons complicates training. Two common remedies exist: converting a trained ANN into an SNN~\cite{rathi2020enabling}, and surrogate gradient learning~\cite{liu2022human}, which estimates the gradient of spike generation during backpropagation. The conversion-based approach underperforms the original ANN and requires large simulation time windows, increasing energy cost and inference latency~\cite{rathi2020enabling}, whereas surrogate gradient learning is more popular because it generally yields better performance~\cite{feng2024brainqn}.

Second, binary spiking neurons encode information using only the values $\{0, 1\}$, which inherently limits their representation capacity. This is the more severe of the two limitations, causing SNNs to perform poorly in complex, high-dimensional environments~\cite{tang2021deep}. Prior mitigations focus on richer spike encoding~\cite{tang2021deep}, but their high computational cost and the need to place the encoding layer off-chip make them inefficient.

Recently, a ternary spiking neuron model that can produce three values, $\{-1, 0, 1\}$, has been introduced to improve the limited representation capacity of the spiking neuron~\cite{guo2024ternary}. This model can describe more complex neural behaviors, such as the dynamic interaction of inhibitory (-1) and excitatory (+1) inputs within the SNN, and has been used to build SNNs for computer vision and natural language processing tasks with strong results~\cite{guo2024ternary,xing2024spikelm}. The SNN requires 2-bit activations per neuron that incur slightly more processing (and energy) overhead. However, no additional multiplication operations are introduced.  

We present the first evaluation of ternary neuron-based SNNs within the deep Q-learning framework. Surprisingly, we find that despite their greater representation capacity, existing ternary spiking neurons perform substantially \emph{worse} than binary ones on deep Q-learning tasks—contradicting the gains they offer in computer vision and language modeling. This counterintuitive degradation drives our central question: why does increased representation capacity fail to translate into better reinforcement learning performance, and how can it be recovered? Our key contributions and findings are summarized as follows.

\begin{itemize}
    \item We mathematically analyze the representation capacity and training dynamics of ternary SNNs, revealing a trade-off between maximizing representation capacity and minimizing gradient estimation bias during training.

    \item We propose a new ternary spiking neuron model with asymmetric thresholds for positive and negative spikes that reduces gradient estimation bias. This bias is especially harmful in RL, where the quality of the replay-buffer training data depends on the agent's own training dynamics; our design is further inspired by the asymmetry of biological neurons.

    \item We show that DSQNs built from our proposed model—which we call \emph{Deep Asymmetric Ternary Spiking Q-Networks} (\emph{DATSQN})—train stably. Using the concept of dynamic isometry~\cite{xing2024spikelm, chen2020comprehensive}, we establish that they avoid both the vanishing and exploding gradient problems.
\end{itemize}

DATSQN is evaluated on seven Atari games from the Gym environment, using the binary spiking neuron-based DSQN architecture from~\cite{liu2022human} as a baseline. To enable fully spiking computation compatible with neuromorphic hardware, we encode raw pixel input into spike trains using rate-based Bernoulli sampling, unlike~\cite{liu2022human}, which processes raw pixels directly, resulting in a non-spiking first layer with floating-point operations. To further challenge the agents while reducing latency and energy consumption, we also use a shorter simulation time window than the baseline. Under these conditions, DATSQN achieves a $30\%$ improvement in mean game score over the binary DSQN baseline.

The paper is organized as follows. Section~\ref{sec:related_work} discusses related work. Section~\ref{sec:preliminary} provides background on deep Q-learning networks and their spiking variants. Section~\ref{sec:method} describes the methodology that led to the development of DATSQN. Section~\ref{sec:results} presents the key results, and we conclude the paper in Section~\ref{sec:conclusion}.

\section{Related Work}\label{sec:related_work}
DRL combines reinforcement learning with deep neural networks; the original deep Q-network (DQN) couples Q-learning with a DNN~\cite{mnih2013playing}. It has since been improved with ideas such as a replay buffer and the $\epsilon$-greedy method~\cite{mnih2015human}. Double DQN addresses overestimation~\cite{van2016deep}, dueling DQN improves the stability of the base network~\cite{wang2016dueling}, and the two have been integrated~\cite{sewak2019deep}.

The high energy consumption of DRL algorithms motivated the first DSQN, which converts a pre-trained DQN into an SNN~\cite{patel2019improved} but yields lower performance than the DQN. An improved conversion approach requires a larger simulation time window, increasing latency and energy consumption~\cite{tan2021strategy}. To mitigate these issues, training of DSQNs using surrogate gradients was developed~\cite{liu2022human}. A batch normalization approach for DSQN improves performance at the cost of additional computation~\cite{sun2022solving}. The impact of hyperparameter tuning on DSQN performance has also been studied~\cite{akl2023toward,feng2024brainqn}.

Although these efforts have advanced DSQN, its effectiveness on complex, high-dimensional tasks is constrained by the limited representation capacity of binary spiking neurons---a limitation that worsens when smaller simulation-time windows are used to reduce latency. While SNNs were first applied to computer vision~\cite{tan2021strategy, sun2022solving}, their more recent use in language modeling~\cite{xing2024spikelm} and reinforcement learning~\cite{feng2024brainqn, lv2019path} has drawn increased attention to this limitation, motivating the ternary spiking neuron model~\cite{guo2024ternary, xing2024spikelm}.

Building on this line of work, we explore the practicality of ternary spiking neurons for enhancing the representation capacity of DSQNs. We find that existing ternary neurons in fact degrade DSQN performance relative to binary ones, and propose a new asymmetric ternary spiking model that resolves this issue while ensuring training stability.

\section{Preliminaries}\label{sec:preliminary}
Consider a task in which an agent interacts with an environment through a sequence of actions, observations, and rewards. At each time step $t$, the agent selects an action from a set of possible actions $\mathcal{A} = \{a_1, \dots, a_k\}$ and applies it to the environment. The agent receives a reward $r_t$ and observes the new state of the environment $s_t \in \mathcal{S}$. The underlying process is a Markov decision process, which means that the state at time $t$ depends on the state $s_{t-1}$ and the action $ a_{t-1} $ taken by the agent at the previous time step. Assuming rewards are discounted by a factor of $ \gamma $ per time step, the cumulative discounted reward (return) at time $t$ is $R_t = \sum_{i=t}^T \gamma^{i-t} r_i$, where $T$ is the time step at which the interaction terminates. The optimal action-value function is the maximum return achievable by any policy and is given as
\begin{equation}
\label{eq:3}
Q^*(s, a) = \mathbb{E} \left[ r + \gamma \max_{a' \in \mathcal{A}} Q^*(s', a') \mid s, a \right],
\end{equation}
where $s'$ is the state reached after the agent takes action $a$. The idea of DQN is to approximate $Q^*(s, a)$ using a deep neural network. Due to the correlations between the action-value function, $Q^*(s, a)$, and the target values, $r + \gamma \max_{a' \in \mathcal{A}} Q(s', a')$, the learning algorithm becomes unstable. Mnih et al. propose an iterative update algorithm which adjusts the action-value function toward target values that are only periodically updated, resulting in a stable training process~\cite{mnih2015human}. This algorithm has been adopted in several subsequent advances in DQN~\cite{zhang2024sf, lv2019path, hester2018deep}.

A DSQN can be established by substituting neurons in a DQN with spiking neurons, such as the LIF or Izhikevich neuron models. These neurons encode information within the network into spike sequences. Owing to their simpler hardware implementation, LIF neurons are widely used in the SNN literature. LIF neurons encode information as a sequence of $0$ or $1$ over a simulation time window, and we refer to them as binary LIF neurons in this work. 
\vspace{3pt}

\textbf{Binary LIF Neuron Model.} The dynamics, described by Equation~\eqref{eq:4}, simulate the charging and firing of biological neurons. Here, $t \in \mathcal{T}$ denotes a time step in the simulation window $\mathcal{T}$, whose length is $|\mathcal{T}|$, over which the binary LIF neuron performs spike encoding; $m^l(t)$ and $v^l(t)$ are the membrane potentials of the neuron in layer $l$ before and after a possible spike emission, respectively, and $v^{\text{th}}$, $\beta$, and $v_{\text{reset}}$ represent the firing threshold, the decay factor, and the reset potential of the membrane, respectively. The neuron in layer $l$ receives the input from the previous layer, $x^{(l-1)}(t)$, and its membrane potential updates to $m^l(t)$. Then, if the updated membrane potential, $m^l(t)$, is greater than $v^{\text{th}}$, the output of the neuron is 1 (spike); otherwise, the output is 0 (no spike). If a spike is generated, the membrane potential will reset to the predefined $v_{\text{reset}}$; otherwise, the membrane potential will decay by a factor denoted by $\beta$. 
\begin{align}\label{eq:4}
m^l(t) &= v^l(t-1) + x^{l-1}(t) \nonumber \\
s^l(t) &= 
\begin{cases} 
0 & \text{if } m^l(t) < v^{\text{th}} \\
1 & \text{if } m^l(t) \geq v^{\text{th}}
\end{cases} \\
v^l(t) &= \beta m^l(t) (1 - s^l(t)) + v_{\text{reset}} s^l(t) \nonumber
\end{align}
In the above encoding mechanism, binary LIF neurons ignore all membrane potentials with negative values, which may still carry valuable information. This limitation makes them less suitable for complex and high-dimensional tasks. To address this challenge, the ternary LIF neuron model was proposed~\cite{guo2024ternary,xing2024spikelm}.
\vspace{3pt} 

\textbf{Ternary LIF Neuron Model.} Spike encoding for the ternary LIF neuron is described by Equation~\eqref{eq:5}. Both negative and positive membrane potentials are considered during encoding, which increases the representation capacity of the neurons. Furthermore, since the ternary LIF encodes the membrane potential in values $\{-1, 0, 1\}$, it still preserves the multiplication-free property of the SNNs, which is critical in maintaining energy efficiency.
\begin{equation}\label{eq:5}
s^l(t) = 
\begin{cases} 
1 & \text{if } m^l(t) \geq v^{\text{th}} \\
0 & \text{if } -v^{\text{th}} < m^l(t) < v^{\text{th}} \\
-1 & \text{if } m^l(t) \leq -v^{\text{th}}
\end{cases}
\end{equation}
Although the ternary neuron improves representation capacity, our experiments on Q-learning tasks show that it performs worse than the binary LIF (see Table~\ref{table4} in Section~\ref{sec:results}). This is doubly surprising: it contradicts both the gains ternary neurons offer in other domains, such as language modeling and computer vision~\cite{guo2024ternary, xing2024spikelm}, and the analysis of Guo et al. showing that ternary LIF increases information entropy~\cite{guo2024ternary}. We would expect DSQN to perform better with increased representation capacity, whereas our observations indicate otherwise. This motivates us to develop a mathematical framework to resolve this contradiction and address both the performance degradation of ternary LIF neurons and the limited representation capacity of binary LIF neurons in Q-learning tasks.

\section{Methodology}\label{sec:method}
We first mathematically analyze spike encoding in SNNs and develop a hypothesis to explain the performance degradation observed when using ternary LIF neurons in DSQN. Building on the insights from this framework, we propose a novel ternary LIF model to overcome the limited representation capacity of the binary LIF. Finally, recognizing the importance of training stability in RL tasks, we mathematically prove that the proposed ternary LIF model ensures stable training dynamics by avoiding the vanishing and exploding gradient problems.

\subsection{Information Loss and Training Performance}\label{subsec:infoloss}
Following Guo et al. and Xing et al., we model spikes as random variables rather than deterministic values~\cite{guo2024ternary, xing2024spikelm}. This lets us describe their behavior through the probability of taking specific values and the distribution of the membrane potential, as defined by the following equations:

\begin{equation}\label{eq:8}
\tilde{s}^B =
\begin{cases}
0 & p^0 = P(m(t) < v^{\text{th}}) \\
1 & p^+ = P(m(t) \geq v^{\text{th}}),
\end{cases}
\end{equation}

\begin{equation}\label{eq:9}
\tilde{s}^T =
\begin{cases}
1 & p^+ = P(m(t) \geq v^{\text{th}}) \\
0 & p^0 = P(-v^{\text{th}} < m(t) < v^{\text{th}}) \\
-1 & p^- = P(m(t) \leq -v^{\text{th}})
\end{cases}
\end{equation}
In the above, $\tilde{s}^B$ and $\tilde{s}^T$ denote stochastic binary and ternary spikes, respectively, and $p^0$, $p^+$, and $p^-$ are the probabilities of observing a zero spike, a positive spike, and a negative spike, respectively. These probabilities are determined by the distribution of the membrane potential $m(t)$. Under conditions that generally hold, the membrane potential of LIF neurons in the subthreshold regime tends to follow a Gaussian-like distribution (see Appendix A). Therefore, we can define $p^0$, $p^+$, and $p^-$ based on the distance of $m(t)$ to the thresholds and/or $0$. As a result, for the binary case, we can write $p^0 + p^+ = 1$, and for the ternary case $p^0 + p^- + p^+ = 1$.
\vspace{3pt}

\textbf{Information Entropy Analysis.}
Consider a set of random variables $\mathcal{V}$. The representation capacity of $\mathcal{V}$ can be quantified using its information entropy $\mathcal{H}(\mathcal{V}) = -\sum_{s \in \mathcal{V}} P(s) \log P(s)$, where $P(s)$ is the probability of observing $s \in \mathcal{V}$. $\mathcal{H}(\mathcal{V})$ reaches its maximum when all outcomes $s \in \mathcal{V}$ are equally likely, that is, $P(s)$ is uniform.

By assuming the same firing rate, $r$, we can use the concept of information entropy to compare the representation capacity of binary and ternary LIFs. For binary LIF, we have: $\mathcal{V} = \{0, 1\}$, $p^0 = 1 - r$, and $p^+ = r$. For ternary LIF neurons, since the threshold $v^{th}$ is symmetric for both the positive and negative domains, and the membrane potential $m(t)$ follows a Gaussian-like distribution, the probabilities of generating positive and negative spikes are equal. Thus, we have $\mathcal{V} = \{-1, 0, 1\}$, $p^0 = 1 - r$, and $p^- = p^+ = \frac{r}{2}$.
The information entropy of binary and ternary LIFs is calculated as
\begin{align}\label{eq:entropy}
\mathcal{H}(\tilde{s}^B) &= -r \ln r - (1 - r) \ln (1 - r) \nonumber \\
\mathcal{H}(\tilde{s}^T) &= -r \ln \left( \frac{r}{2} \right) - (1 - r) \ln (1 - r)
\end{align}
We have $\mathcal{H}(\tilde{s}^T)-\mathcal{H}(\tilde{s}^B)=r \ln 2$ nats. Thus, ternary LIF increases the representation capacity of spiking neurons by $r$ bits.
\vspace{3pt}

\textbf{Gradient Within the Network.}
Although spike encoding is inherently non-differentiable due to the use of a threshold function (see Equation \eqref{eq:5}), methods such as surrogate gradient learning (SGL)~\cite{liu2022human,41,42} and the straight-through estimator (STE)~\cite{chen2020comprehensive} offer effective solutions for training SNNs via backpropagation by estimating gradients. Considering stochastic spike encoding introduced in Equations~\eqref{eq:8} and \eqref{eq:9}, we can calculate the expected value of the spike-generation gradient to facilitate the analysis of backpropagation through the network. This is given by
\begin{align}
\mathbb{E}\left[\frac{\partial \tilde{s}^B}{\partial m(t)}\right] &= \frac{\partial}{\partial m(t)} \mathbb{E}[\tilde{s}^B] = \frac{\partial}{\partial m(t)} \left( 0 \cdot p^0 + 1 \cdot p^+ \right) \label{eq:10} \\
\mathbb{E}\left[\frac{\partial \tilde{s}^T}{\partial m(t)}\right] &= \frac{\partial}{\partial m(t)} \mathbb{E}[\tilde{s}^T] = \frac{\partial}{\partial m(t)} \left( 0 \cdot p^0 - 1 \cdot p^- + 1 \cdot p^+ \right) \label{eq:11}
\end{align}

Equations~\eqref{eq:10} and~\eqref{eq:11} give the gradient of the expected \emph{output}. For a stochastic spiking unit, however, the signal that actually drives backpropagation is the gradient of the expected \emph{loss}. Let $L(k)$ denote the loss the network incurs when the neuron emits spike value $k \in \{-1, 0, 1\}$ (the rest of the network marginalized out), and let $J = \mathbb{E}[L(\tilde{s}^T)] = \sum_k p^k L(k)$ be the expected loss. By the score-function identity,
\begin{equation}\label{eq:policy}
\frac{\partial J}{\partial m(t)} = \sum_{k \in \{-1,0,1\}} L(k)\,\frac{\partial p^k}{\partial m(t)}.
\end{equation}
Because the membrane potential follows a Gaussian-like distribution (Appendix~A), the class sensitivities are
\begin{equation}\label{eq:sens}
\frac{\partial p^+}{\partial m(t)} = c_p, \qquad
\frac{\partial p^-}{\partial m(t)} = -c_n, \qquad
\frac{\partial p^0}{\partial m(t)} = c_n - c_p,
\end{equation}
with $c_p, c_n > 0$ the positive- and negative-threshold sensitivities (equal, $c_p = c_n = c$, for the symmetric thresholds of Equation~\eqref{eq:5}): raising the membrane potential creates positive spikes and removes negative spikes. Substituting~\eqref{eq:sens} into~\eqref{eq:policy} and writing the loss through a \emph{sign} component $g = \tfrac{1}{2}[L(1) - L(-1)]$ and an \emph{activity} component $e = \tfrac{1}{2}[L(1) + L(-1)] - L(0)$ yields
\begin{equation}\label{eq:master}
\frac{\partial J}{\partial m(t)}
= \underbrace{(c_p + c_n)\,g}_{\text{sign channel}}
+ \underbrace{(c_p - c_n)\,e}_{\text{activity channel}}.
\end{equation}
The expected gradient splits into a sign channel, carrying the loss's preference between $+1$ and $-1$, and an activity channel, carrying its preference between firing and silence. For the symmetric ternary LIF, $c_p = c_n$, so the activity channel vanishes identically and
\begin{equation}\label{eq:symgrad}
\frac{\partial J}{\partial m(t)} = 2c\,g = c\,[L(1) - L(-1)].
\end{equation}
The symmetric neuron can thus be taught only which sign to prefer, never whether to be active, and even this surviving signal vanishes whenever the loss does not discriminate spike sign, $L(1) = L(-1)$. Common estimators such as STE share this blindness: averaged over the membrane distribution, they reduce to the same sign channel $2cg$, so the activity channel is invisible to them as well.

In RL, the sign component $g$ is the network's current estimate of whether driving a feature positive or negative reduces the temporal-difference error. Early in training, the bootstrapped targets are uninformative about this polarity, so $g \approx 0$ on average while its variance is large: the symmetric ternary neuron then receives a near-zero, high-variance expected gradient and stalls, even when activity strongly affects the loss ($e \neq 0$). In computer vision and language modeling, the labels fix the sign preference from the outset, so $g$ is nonzero and the stall does not arise, which is why ternary neurons help there yet degrade DSQN, as summarized in the following Lemma.

\begin{lemma}
\label{lem:lemma1}
Consider an SNN using the symmetric ternary LIF of Equation~\eqref{eq:5}. At the maximum-capacity operating point $p^+ = p^-$, the expected gradient of the loss reduces to the sign channel $2cg$ of Equation~\eqref{eq:symgrad} and is independent of the activity channel $e$. It vanishes whenever the loss does not discriminate spike sign ($L(1) = L(-1)$), as holds for the uninformative targets of early Q-learning.
\end{lemma}
\vspace{-12pt}
\begin{proof}
As established in the entropy analysis above, capacity is maximized when $p^+ = p^-$, where the thresholds are symmetric and $c_p = c_n$. The activity term of Equation~\eqref{eq:master} then drops, leaving $2cg$. This is zero when $g = \tfrac{1}{2}[L(1) - L(-1)] = 0$, and by the chain rule, the expected gradient is correspondingly zero.
\end{proof}

\begin{remark} 4.1.
\label{rem:rem1}
Because the symmetric neuron's expected gradient carries only the sign channel, it is blind to how activity affects the loss; in the sign-undetermined regime of early RL its expected gradient is near zero with large variance, so STE and related estimators are highly biased. This adversely affects training and accounts for the degradation of symmetric ternary LIF relative to binary LIF in Q-learning tasks.
\end{remark}

\subsection{Asymmetric Ternary LIF Neuron}
Lemma~\ref{lem:lemma1} and Remark~\ref{rem:rem1} reveal a tension: maximizing representation capacity forces the thresholds to be symmetric, $c_p = c_n$, which closes the activity channel $(c_p - c_n)e$ of Equation~\eqref{eq:master} and leaves a learning signal that vanishes whenever spike sign is undetermined. We therefore break the threshold symmetry---raising capacity above the binary neuron while keeping $c_p \neq c_n$, so the activity channel stays open. Given that the membrane potential follows a Gaussian-like distribution in the subthreshold region, symmetric thresholds make $p^+ = p^-$ likely; assigning distinct positive and negative thresholds keeps the two branch sensitivities unequal and the expected gradient nonzero. We propose an asymmetric ternary spiking model, as follows:
\begin{equation}\label{eq:15}
s^l(t) =
\begin{cases}
1 & \text{if } m^l(t) \geq v^{\text{th}_p} \\
0 & \text{if } -v^{\text{th}_n} < m^l(t) < v^{\text{th}_p} \\
-1 & \text{if } m^l(t) \leq -v^{\text{th}_n}
\end{cases}
\end{equation}
where $v^{\text{th}_p}$ and $v^{\text{th}_n}$ are the positive and negative firing thresholds, respectively, and $v^{\text{th}_p} \neq v^{\text{th}_n}$.

We fix $v^{\text{th}_p}$ at a constant value and treat $v^{\text{th}_n}$ as a trainable parameter, which enables the optimizer to adjust $v^{\text{th}_n}$ together with other trainable parameters, allowing the network to implicitly influence the bias of the chosen gradient estimator. In Theorem \ref{thm:th1}, we prove that the proposed asymmetric ternary model enhances information entropy and prevents the expected value of the gradient from becoming zero.

\begin{theorem}
\label{thm:th1}
Consider the binary and asymmetric ternary spiking neuron models described in Equations \eqref{eq:4} and \eqref{eq:15}, respectively. For the same firing rate $r$, the asymmetric ternary model achieves higher representation capacity and ensures that the expected value of its gradient remains nonzero.
\end{theorem}

\vspace{-12pt}

\begin{proof}
The representation capacity can be quantified by the information entropy. For binary LIF, $\mathcal{V} = \{0, 1\}$, $p^0 = 1 - r$, and $p^+ = r$. For asymmetric ternary LIF, given that $v^{\text{th}_n} \neq v^{\text{th}_p}$ in general, we can write: $\mathcal{V} = \{-1, 0, 1\}$, $p^0 = 1 - r$, $p^- + p^+ = r$. Assume that $p^- = r_1$ and $p^+ = r_2$. We can calculate the information entropy of the binary LIF and the asymmetric ternary LIF as
\begin{align}\label{eq:entropyat}
\mathcal{H}(\tilde{s}^B) &= -r \ln r - (1 - r) \ln(1 - r) \nonumber \\
\mathcal{H}(\tilde{s}^{AT}) &= -r_1 \ln r_1 - r_2 \ln r_2 - (1 - r) \ln(1 - r)
\end{align}
Since $r = r_1 + r_2$, $\mathcal{H}(\tilde{s}^{AT}) - \mathcal{H}(\tilde{s}^B) = -r_1 \ln \left( \frac{r_1}{r} \right) - r_2 \ln \left( \frac{r_2}{r} \right)$ and since $r_1, r_2 > 0$, we can conclude that
$-r_1 \ln \left( \frac{r_1}{r} \right) - r_2 \ln \left( \frac{r_2}{r} \right) > 0$,
which means an increase in the information entropy, leading to higher representation capacity of the asymmetric ternary LIF compared to the binary LIF. 
The expected gradient of the asymmetric ternary LIF follows from the master equation~\eqref{eq:master}: since $v^{\text{th}_n} \neq v^{\text{th}_p}$ gives $c_p \neq c_n$, the activity channel $(c_p - c_n)e$ is nonzero, so the expected gradient remains nonzero even when the sign channel vanishes ($g = 0$), and the proof is completed.
\end{proof}

\subsection{Analysis of Training Stability}
Given the importance of training dynamics in RL, we analyze SNNs with asymmetric ternary spiking neurons through the lens of block dynamic isometry~\cite{xing2024spikelm, chen2020comprehensive}, focusing on their susceptibility to the exploding or vanishing gradient problem. Consider a neural network as a series of blocks $f(x) = f_{\theta^L}^L f_{\theta^{(L-1)}}^{(L-1)} \cdots f_{\theta^1}^1 (x)$, where \( f_{\theta^j}^j \) represents the \( j \)-th layer of the network, and \( \theta^j \) denotes the trainable parameters of layer \( j \). The Jacobian matrix for block \( j \) is defined as \( J_j = \frac{\partial f^j}{\partial f^{j-1}} \). If we define $\phi(J) = \mathbb{E}[\text{tr}(J)]$, then $\varphi(J) = \phi(J^2) - \phi(J)^2$.

\begin{definition}
\label{def:def1}
Consider a neural network that can be represented as a sequence of blocks. \( J_j \) denotes the \( j \)-th block’s Jacobian matrix. If \( \forall j, \phi(J_j J_j^T) \approx 1 \), and \( \varphi(J_j J_j^T) \approx 0 \), then the network achieves block dynamic isometry~\cite{chen2020comprehensive}.
\end{definition}

A network that achieves dynamic isometry avoids vanishing or exploding gradients during training by maintaining all values of its input-output Jacobian matrix $\approx 1$. We analyze the dynamic isometry of networks composed of asymmetric spiking neurons. Given the widespread success of the ReLU activation function in enabling stable training across various machine learning tasks, we compare the block isometry of networks using asymmetric ternary neurons to those employing ReLU activations.

\begin{lemma}
    \label{lem:lem2}
    If the probability of the input being greater than $0$ is $p$ and the network uses ReLU functions, then $\phi(J_j J_j^T )=p$ and $\varphi(J_j J_j^T )=p-p^2$~\cite{chen2020comprehensive}.
\end{lemma}

For spiking neural networks, let $s_m$ denote the Jacobian of the threshold function in Equation~\eqref{eq:15}; due to its element-wise nature, $s_m$ is diagonal.

\begin{lemma}
    \label{lem:lem3}
    If the spike firing rate of neurons is $r$, and the network uses asymmetric ternary spiking neurons as described in Equation \eqref{eq:15}, then $\phi(s_m s_m^T )=1-r$ and $\varphi(s_m s_m^T )=r-r^2$.
\end{lemma}

\begin{theorem}
    \label{thm:th2}
    The asymmetric ternary spiking neuron achieves at least the dynamic isometry of ReLU activations, that is, $\phi(s_m s_m^T ) \geq \phi(J_j J_j^T )$ and $\varphi(s_m s_m^T ) \leq \varphi(J_j J_j^T )$.
\end{theorem}

Theorem \ref{thm:th2} ensures the training stability of SNNs using our proposed neuron model.

\textbf{Biological Interpretation.} Biology classifies neurons as excitatory or inhibitory, with about four out of every five neurons being excitatory. Research indicates that when 70--80\% neurons are excitatory, the brain's ability to solve complex problems is maximized~\cite{wang2024we}. The ternary spiking neuron model allows neurons to exhibit both excitatory (emitting \(+1\)) and inhibitory (emitting \(-1\)) behaviors. The basic ternary LIF model described in Equation \eqref{eq:5} makes the neuron respond symmetrically: its inhibitory effect for negative inputs mirrors its excitatory effect for positive inputs. However, this symmetry contradicts biological evidence. In contrast, the asymmetric ternary spiking neuron model described in Equation \eqref{eq:15} enables neurons to exhibit a dominance of excitatory or inhibitory behavior, aligning more closely with the brain.

\section{Experiments}\label{sec:results}
We first examine our hypothesis that gradient estimation bias is a key factor contributing to the observed performance degradation of ternary LIF neurons. Motivated by this hypothesis, we then evaluate how the increased representation capacity of the asymmetric ternary neuron affects SNN performance on Q-learning tasks.

\textbf{Experimental Setup}. Seven games from the OpenAI Gym Atari environment were selected for our evaluation; the set was fixed in advance, before observing any results, to avoid selection bias. Liu et al. demonstrated that SNNs using binary spiking neurons can achieve human-level performance on Atari tasks~\cite{liu2022human}. However, their approach relies on a simulation time window of $40$, leading to a long decision-making latency. To make the task more challenging while reducing latency, our experiments use a shorter simulation time window of $20$. When discussing the results, we refer to binary spiking neuron-based DQN, ternary spiking neuron-based DQN, and asymmetric ternary spiking neuron-based DQN as DSQN, DTSQN, and DATSQN, respectively.\footnote{The source code is available at: https://github.com/Aref7792/DATSQN.}.

\textbf{Network Architecture}. We adopt the architecture used by Liu et al.~\cite{liu2022human} (and originally proposed by Mnih et al.~\cite{mnih2015human}), consisting of three convolutional layers followed by two fully connected layers. The first fully connected layer contains 512 neurons, while the second layer has between 4 and 18 neurons, depending on the number of possible actions in each game. All neurons in the network are spiking neurons of the same type—binary, ternary, or asymmetric ternary, depending on the selected neuron model.

\textbf{Input Encoding}. To enable a fully spiking network—consistent with the principles of neuromorphic hardware—we apply rate-based encoding using Bernoulli sampling to convert raw pixel values into spike trains. This ensures compatibility with spike-based computation and reduces the need for costly floating-point operations.

\begin{figure*}[t!]
    \centering
{\includegraphics[width=0.88\textwidth]{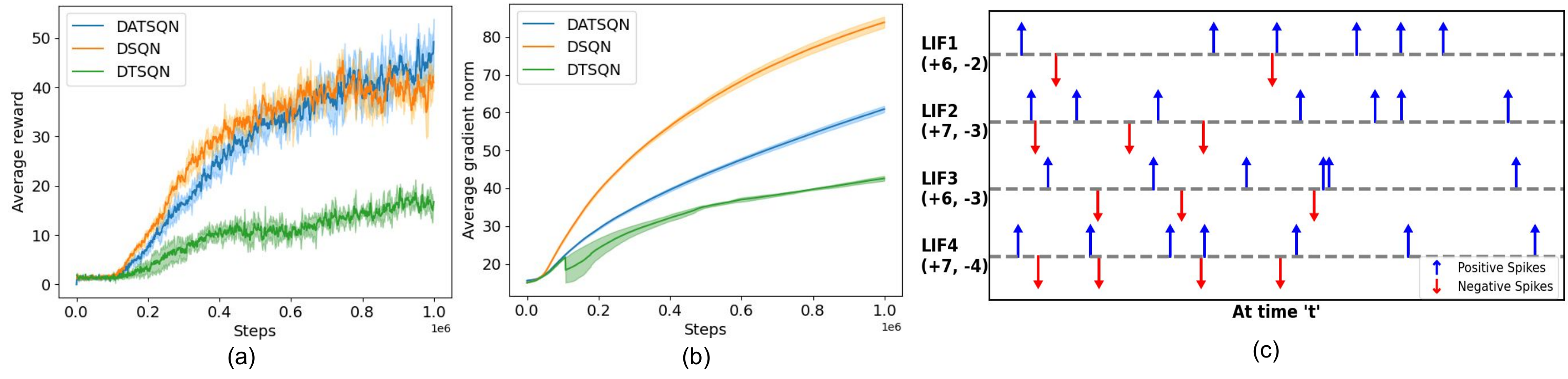}}
    \caption{(a) Learning curves, (b) average gradient norm across the network during training for each RL agent, and (c) excitatory and inhibitory spikes generated by randomly selected neurons in the breakout environment. Shaded bands in (a) and (b) denote standard deviation across the five seeds.}
    \label{fig:fig1}
\end{figure*}

\begin{figure*}[t!]
    \centering
    {\includegraphics[width=.63\textwidth]{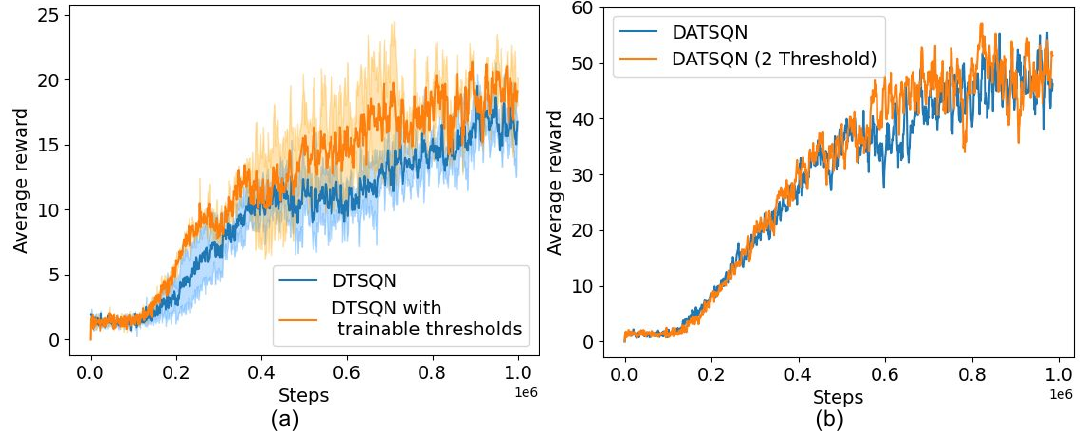}}

    \vspace{-0.05cm}
    {\includegraphics[width=.63\textwidth]{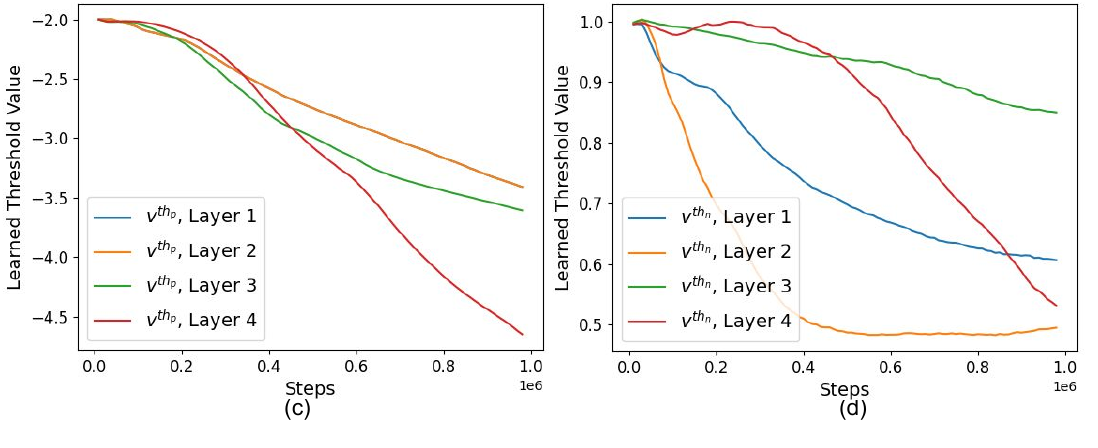}}
    \vspace{-10pt}
    \caption{Learning curves for (a) DTSQN with and without trainable thresholds, and (b) DATSQN with one (canonical) versus two trainable threshold parameters in the Breakout environment. Dynamic evolution of the negative thresholds (c) and the positive thresholds (d) in asymmetric ternary spiking neurons within the hidden layers of the two-trainable-threshold DATSQN variant. Shaded bands in (a) and (b) denote standard deviation across the five seeds.}
    \label{fig:ablation1}
    \vspace{-12pt}
\end{figure*}

\textbf{Testing and Metrics}. We train each agent from scratch with an $\epsilon$-greedy algorithm ($\epsilon = 0.1$) over $10^6$ steps, using five random seeds shared across DSQN, DTSQN, and DATSQN to enable a paired comparison. For each seed, we checkpoint the policy every 10{,}000 steps, select the best checkpoint by evaluation reward, and evaluate it over 10 episodes. All reported scores and error bars---in Table~\ref{table4} and as the shaded bands in Figs.~\ref{fig:fig1}(a,b) and~\ref{fig:ablation1}(a,b)---are the mean and standard deviation across the five seeds. To facilitate comparison, we also normalize the scores of DTSQN and DATSQN with respect to the DSQN score. Complete implementation details and all hyperparameters are provided in our code repository.

\begin{table}[t!]
\centering
\small

\begin{minipage}{0.99\columnwidth} 
\fontsize{7.5}{10}\selectfont
    \centering
    \caption{Scores for DSQN, DTSQN, and DATSQN (mean $\pm$ std across five seeds). Bold: DATSQN exceeds DSQN.}
    \label{table4}
    \begingroup
    \setlength{\tabcolsep}{5.5pt} 
    \renewcommand{\arraystretch}{.85} 
    \begin{tabular}{lccc}
        \toprule
        Game & DSQN & DTSQN & DATSQN \\
        & score $\pm$ (std)  & score $\pm$ (std)  & score $\pm$ (std)  \\
        \midrule
        Beam Rider & $2069 \pm (580.4)$ & $1518.6 \pm (529.7)$ & $\mathbf{4000 \pm (967.25)}$ \\
        Boxing & $79.3 \pm (9.5)$ & $49.1 \pm (18.5)$ & $79.1 \pm (7.1)$\\
        Breakout  & $207.7 \pm (64)$ & $23.6 \pm (6.5)$ & $\mathbf{252.14 \pm (70)}$ \\
        Crazy Climber & $53230 \pm (11660)$ & $4240 \pm (1110)$ & $\mathbf{54190 \pm (13921)}$ \\
        Gopher & $1842 \pm (465)$ & $1162 \pm (564.06)$ & $\mathbf{2111.1 \pm (648)}$ \\
        Jamesbond & $185 \pm (109)$ & $120 \pm (52)$ & $\mathbf{250 \pm (140)}$ \\
        SpaceInvaders & $457 \pm (64.8)$ & $324 \pm (102.3)$ & $\mathbf{655 \pm (123.7)}$ \\
        \bottomrule
    \end{tabular}
    \endgroup
\end{minipage}
\vspace{-12pt}
\end{table}

\textbf{Performance Analysis}. Table~\ref{table4} reports the scores for three RL agents in seven games. DATSQN achieves higher scores than DSQN in six of the seven games, with a mean relative improvement of 30\% (median 21\%), while DTSQN shows a clear performance drop in every game. These results are consistent with our hypothesis because the asymmetric design addresses the performance issues of ternary neurons in Q-learning. The improvement of DATSQN over DSQN likely comes from retaining the higher representation capacity of ternary neurons while restoring an informative gradient signal---the two properties that a symmetric ternary neuron cannot satisfy at once. We also note that our DSQN scores are lower than those reported in~\cite{liu2022human}, for two reasons: their first layer is non-spiking, relying on floating-point computation that is less compatible with neuromorphic hardware; and we use a shorter simulation time window to reduce latency, making the task more challenging.

\textbf{Training Dynamics}. Figure~\ref{fig:fig1} shows the average cumulative reward during training and the average gradient norm for the three agents in the Breakout environment. As Fig.~\ref{fig:fig1} (a) indicates, DTSQN consistently shows a degradation in performance compared to the state-of-the-art DSQN. Furthermore, based on Fig.~\ref{fig:fig1} (b), DTSQN has the lowest average gradient norm among all three agents, empirically reflecting the weakened learning signal that Lemma~\ref{lem:lemma1} predicts for symmetric ternary neurons. During the initial training steps, DSQN shows better performance than DATSQN in terms of the cumulative reward achieved. However, during the final stages of training, DATSQN achieves a higher cumulative reward (sign of increased representation capacity). The average gradient norm during training remains stable, showing no sudden spikes or sharp drops, indicating no signs of vanishing or exploding gradients. 

To better understand learning behavior, we randomly select neurons with asymmetric domains from each layer of various SNN models and, within a randomly chosen feature map of a convolutional layer, track the positive and negative spikes reaching each neuron from the $( k_h \times k_w )$ region of the previous layer's feature map. This mirrors a convolution over the $(k_h \times k_w)$ region. Empirical observations in all models and layers consistently show that the number of positive spikes exceeds the number of negative spikes, with the ratio $\ge 60\%$. Figure~\ref{fig:fig1}(c) illustrates this for four neurons in the second convolutional layer (a $4 \times 4$ region).

Positive spikes act as excitatory impulses, whereas negative spikes are inhibitory. Our SNN model that uses ternary neurons with an asymmetric domain is consistent with a fundamental property of the human brain, where approximately 70--80\% of neurons are excitatory and the remainder are inhibitory~\cite{sultan2018generation, wang2024we}. For DRL tasks that require adaptive decision making in dynamic environments, we believe that LIF neurons with asymmetric domains can play a critical role in facilitating effective learning by maintaining an optimal excitation-inhibition balance.

\textbf{Ablation Study.} We attribute DATSQN's improved performance to improved gradient estimation quality, which prevents the expected value of the gradient from diminishing toward zero. However, this improvement may also be due to including a trainable threshold in the ternary neuron model. To further investigate this, we performed a study by training DTSQN with the trainable threshold, making \( v^{\text{th}} \) in Equation~\eqref{eq:5} a trainable parameter. The results, shown in Fig.~\ref{fig:ablation1} (a), indicate that introducing a trainable threshold only marginally improves the performance of DTSQN; a substantial performance gap remains relative to the baseline (DSQN). It may be that with only one threshold parameter, the positive and negative thresholds remain symmetric, so by Lemma~\ref{lem:lemma1} the expected gradient collapses to its sign channel and, under the uninformative targets of early Q-learning, provides little learning signal.

In our canonical DATSQN, only the negative threshold \( v^{\text{th}_n} \) is trainable, while the positive threshold \( v^{\text{th}_p} \) is fixed. Our final set of experiments examines whether making both thresholds trainable helps, by evaluating a DATSQN variant in which neurons are equipped with two independent trainable thresholds. The results, shown in Fig.~\ref{fig:ablation1} (b), indicate that this additional flexibility does not improve performance much over the single-trainable-threshold DATSQN, which justifies our design choice. The dynamic evolution of these thresholds during training, for the two-threshold variant, is shown in Fig.~\ref{fig:ablation1} (c) and (d), where the network decreases the absolute value of the positive threshold while increasing that of the negative threshold. Since the network is optimized to maximize cumulative rewards, this behavior suggests that neurons tend to act more excitatory than inhibitory, supporting the biologically inspired motivation discussed in Section~\ref{sec:method}.
\vspace{-12pt}

\section{Conclusion}\label{sec:conclusion}
This paper investigates the use of ternary spiking neurons to improve the performance of SNNs in Q-learning tasks by increasing their representation capacity. Motivated by both mathematical analysis and biological insights, we propose a new ternary spiking neuron model called the asymmetric ternary spiking neuron to address the performance degradation seen in state-of-the-art ternary models. The results show that the proposed model achieves much better performance compared to binary SNNs in Q-learning tasks. These results suggest that the asymmetric ternary neuron is worth evaluating in other SNN settings, such as vision-based tasks and language modeling.


%


\bibliographystyle{IEEEtran}
\bibliography{icons}


\appendix

\appendix

\section{LIF Neuron Dynamics}
Consider an LIF neuron located in the first hidden layer of a large SNN that receives input in the form of spikes from the preceding (input) layer, each modulated by its corresponding synaptic weight. In the subthreshold regime, where the membrane potential remains below the firing threshold, the neuron’s dynamics can be described by the continuous-time version of the LIF model as
\begin{equation}\label{A1}
\tau \frac{dm(t)}{dt} = -\left(m(t) - V_{\text{reset}}\right) + \sum_k \sum_{t_k^i} w_k \, \delta(t - t_k^i),
\end{equation}
where \( \tau \) is the membrane time constant; \( w_k \) denotes the synaptic weight between the target neuron and the \( k \)-th input; \( t_k^i \) represents the time of the \( i \)-th spike from the \( k \)-th input; \( \delta(t) \) is the Dirac delta function; and \( m(t) \) and \( V_{\text{reset}} \) denote the membrane potential and the reset potential, respectively~\cite{gerstner2002spiking}.

Assuming rate-based encoding in the input layer, where raw input values are converted into spike trains via Bernoulli sampling, and that the inputs are independent or only weakly correlated (a condition often satisfied in practice), the Central Limit Theorem implies that the aggregated input to a neuron can be approximated by a Gaussian distribution. To illustrate this in our setting, consider the first convolutional layer of the network, which uses an \(8 \times 8\) kernel and operates on inputs with four channels. We consider a flattened input of size 256, with each element independently drawn from a uniform distribution over \([0, 1]\), mimicking raw pixel intensities. Each input element produces a spike with a probability proportional to its magnitude. For synaptic weights \(w_k\), we use the kernel parameters of one of the best-performing models.  

\begin{figure}[t!]
    \centering
    {\includegraphics[width=.80\columnwidth]{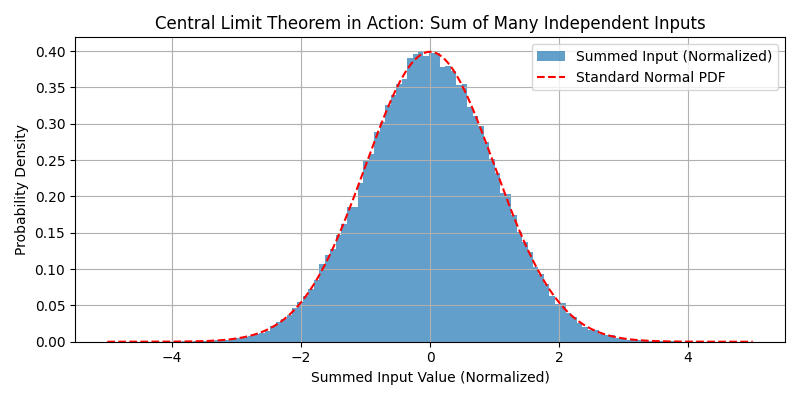}}
    \vspace{-12pt}
    \caption{Approximate Gaussian distribution of the input term \( \sum_k \sum_{t_k^i} w_k \, \delta(t - t_k^i) \), with 500 uniformly distributed samples converted to spike trains via Bernoulli sampling.}
    \label{fig:app1}
    \vspace{-12pt}
\end{figure}

Figure~\ref{fig:app1} shows the resulting distribution of the term \( \sum_k \sum_{t_k^i} w_k \, \delta(t - t_k^i) \), which represents the total weighted input current to a postsynaptic neuron. Thus, Equation~\eqref{A1} can be written as follows: 
\begin{equation}
\tau \frac{dm(t)}{dt} = -\left(m(t) - V_{\text{reset}}\right) + \rho(t),
\label{A2}
\end{equation}
where \( \rho(t) \) is a white noise process that represents the aggregated input \( \sum_k \sum_{t_k^i} w_k \, \delta(t - t_k^i) \). We note that Equation~\eqref{A2} describes the \textit{Ornstein--Uhlenbeck stochastic process} which satisfies the \textit{Fokker--Planck equation} governing the evolution of the probability density function associated with the membrane potential dynamics~\cite{gerstner2002spiking}. This function is given by
\begin{align}
\tau \frac{\partial}{\partial t} p(m, t) = &-\frac{\partial}{\partial m} \left[ \left( -m + V_{\text{reset}} + \tau \sum_k w_k \nu_k(t) \right) p(m, t) \right] + \nonumber \\
& \frac{1}{2} \xi^2 \frac{\partial^2}{\partial m^2} p(m, t),
\label{A3}
\end{align}
where \( p(m, t) \) denotes the probability density of the neuron having membrane potential \( m \) at time \( t \), and \( \nu_k(t) \) represents the spike rate of the \( k \)-th input. The term \( \xi \) is the time-dependent noise amplitude, defined as $\xi^2 = \tau \sum_k w_k \, \nu_k^2(t)$. In the subthreshold regime, assuming \( \mathbb{E}[\rho(t)] \approx 0 \), the expected trajectory of the membrane potential can be obtained by solving Equation~\eqref{A2}, which yields the expression $m_0(t) = \mathbb{E}[m(t)] = V_{\text{reset}} \left(1 - e^{-t / \tau} \right).$
Fluctuations in the membrane potential have a variance \( \sigma_m^2 \), which can be calculated as $\sigma_m^2(t) = \frac{1}{2} \xi^2 \left(1 - e^{-t / \tau} \right)$. Finally, the solution to Equation~\eqref{A3} can be calculated as  
\begin{equation}\label{A4}
p(m, t) = \frac{1}{\sqrt{2\pi \sigma_m^2(t)}} \exp\left( -\frac{(m(t) - m_0(t))^2}{2\sigma_m^2(t)} \right)
\end{equation}
Equation~\eqref{A4} indicates that $p(m,t)$ follows a Gaussian-like distribution centered at $m_0(t)$, or at zero under the common assumption $V_{\text{reset}} = 0$. Although \eqref{A4} characterizes the subthreshold dynamics of an LIF neuron without a firing threshold, it remains a useful approximation when one is present: the threshold acts as an absorbing barrier that, once reached, resets the state and truncates the upper tail, biasing the membrane potential toward $V_{\text{reset}}$ and yielding an approximately truncated Gaussian distribution~\cite{gerstner2002spiking}.

\end{document}